\documentclass{article}

     \usepackage[preprint]{neurips_2022}

\usepackage{graphicx}
\usepackage[utf8]{inputenc} 
\usepackage[T1]{fontenc}    
\usepackage{hyperref}       
\usepackage{url}            
\usepackage{booktabs}       
\usepackage{amsfonts}       
\usepackage{nicefrac}       
\usepackage{microtype}      
\usepackage{xcolor}         

\title{Track Anything: High-performance Interactive Tracking and Segmentation}
\title{Track Anything: High-performance Object Tracking in Videos by Interactive Masks}
\title{Track Anything: Segment Anything Meets Videos}

\author{%
  Jinyu Yang\thanks{Equal contribution. Alphabetical order.},\enskip Mingqi Gao\footnotemark[1],\enskip Zhe Li\footnotemark[1],\enskip Shang Gao, Fangjing Wang, Feng Zheng\thanks{Corresponding author (email: \href{mailto:f.zheng@ieee.org}{f.zheng@ieee.org}).} \\
  SUSTech VIP Lab\\
}

\begin{document}

\maketitle

\begin{abstract}

Recently, the Segment Anything Model (SAM) gains lots of attention rapidly due to its impressive segmentation performance on images.
Regarding its strong ability on image segmentation and high interactivity with different prompts, we found that it performs poorly on consistent segmentation in videos.
Therefore, in this report, we propose Track Anything Model (TAM), which achieves high-performance interactive tracking and segmentation in videos.
To be detailed, given a video sequence, only with very little human participation, \textit{i.e.}, several clicks, people can track anything they are interested in, and get satisfactory results in one-pass inference.
Without additional training, such an interactive design performs impressively on video object tracking and segmentation.
All resources are available on \url{https://github.com/gaomingqi/Track-Anything}.
We hope this work can facilitate related research.

\end{abstract}

\section{Introduction}

Tracking an arbitrary object in generic scenes is important, and Video Object Tracking (VOT) is a fundamental task in computer vision.
Similar to VOT, Video Object Segmentation (VOS) aims to separate the target (region of interest) from the background in a video sequence, which can be seen as a kind of more fine-grained object tracking.
We notice that current state-of-the-art video trackers/segmenters are trained on large-scale manually-annotated datasets and initialized by a bounding box or a segmentation mask.
On the one hand, the massive human labor force is hidden behind huge amounts of labeled data.
Moreover, current initialization settings, especially the semi-supervised VOS, need specific object mask groundtruth for model initialization.
How to liberate researchers from labor-expensive annotation and initialization is much of important.

Recently, Segment-Anything Model (SAM)~\cite{sam} has been proposed, which is a large foundation model for image segmentation.
It supports flexible prompts and computes masks in real-time, thus allowing interactive use.
We conclude that SAM has the following advantages that can assist interactive tracking:
\textbf{1) Strong image segmentation ability.}
Trained on 11 million images and 1.1 billion masks, SAM can produce high-quality masks and do zero-shot segmentation in generic scenarios.
\textbf{2) High interactivity with different kinds of prompts. }
With input user-friendly prompts of points, boxes, or language, SAM can give satisfactory segmentation masks on specific image areas.
However, using SAM in videos directly did not give us an impressive performance due to its deficiency in temporal correspondence.

On the other hand, tracking or segmenting in videos faces challenges from scale variation, target deformation, motion blur, camera motion, similar objects, and so on~\cite{vos,vot6,vot7,vot8,vot9,vot10}.
Even the state-of-the-art models suffer from complex scenarios in the public datasets~\cite{xmem}, not to mention the real-world applications.
Therefore, a question is considered by us:
\textit{can we achieve high-performance tracking/segmentation in videos through the way of interaction?}

In this technical report, we introduce our Track-Anything project, which develops an efficient toolkit for high-performance object tracking and segmentation in videos.
With a user-friendly interface, the Track Anything Model (TAM) can track and segment any objects in a given video with only one-pass inference.
Figure~\ref{fig:overview} shows the one-pass interactive process in the proposed TAM.
In detail, TAM combines SAM~\cite{sam}, a large segmentation model, and XMem~\cite{xmem}, an advanced VOS model.
As shown, we integrate them in an interactive way.
Firstly, users can interactively initialize the SAM, \textit{i.e.}, clicking on the object, to define a target object;
then, XMem is used to give a mask prediction of the object in the next frame according to both temporal and spatial correspondence;
next, SAM is utilized to give a more precise mask description;
during the tracking process, users can pause and correct as soon as they notice tracking failures.

Our contributions can be concluded as follows:

1) We promote the SAM applications to the video level to achieve interactive video object tracking and segmentation.
Rather than separately using SAM per frame, we integrate SAM into the process of temporal correspondence construction.

2) We propose one-pass interactive tracking and segmentation for efficient annotation and a user-friendly tracking interface, which uses very small amounts of human participation to solve extreme difficulties in video object perception.

3) Our proposed method shows superior performance and high usability in complex scenes and has many potential applications.



\section{Track Anything Task}

Inspired by the Segment Anything task~\cite{sam}, we propose the Track Anything task, which aims to flexible object tracking in arbitrary videos.
Here we define that the target objects can be flexibly selected, added, or removed in any way according to the users' interests.
Also, the video length and types can be arbitrary rather than limited to trimmed or natural videos.
With such settings, diverse downstream tasks can be achieved, including single/multiple object tracking, short-/long-term object tracking, unsupervised VOS, semi-supervised VOS, referring VOS, interactive VOS, long-term VOS, and so on.

\section{Methodology}

\subsection{Preliminaries}

\textbf{Segment Anything Model~\cite{sam}.}
Very recently, the Segment Anything Model (SAM) has been proposed by Meta AI Research and gets numerous attention.
As a foundation model for image segmentation, SAM is based on ViT~\cite{vit} and trained on the large-scale dataset SA-1B~\cite{sam}.
Obviously, SAM shows promising segmentation ability on images, especially on zero-shot segmentation tasks.
Unfortunately, SAM only shows superior performance on image segmentation, while it cannot deal with complex video segmentation.

\textbf{XMem~\cite{xmem}.}
Given the mask description of the target object at the first frame, XMem can track the object and generate corresponding masks in the subsequent frames.
Inspired by the Atkinson-Shiffrin memory model, it aims to solve the difficulties in long-term videos with unified feature memory stores.
The drawbacks of XMem are also obvious: 1) as a semi-supervised VOS model, it requires a precise mask to initialize; 2) for long videos, it is difficult for XMem to recover from tracking or segmentation failure.
In this paper, we solve both difficulties by importing interactive tracking with SAM.

\textbf{Interactive Video Object Segmentation.}
Interactive VOS~\cite{mivos} takes user interactions as inputs, \textit{e.g.}, scribbles.
Then, users can iteratively refine the segmentation results until they are satisfied with them.
Interactive VOS gains lots of attention as it is much easier to provide scribbles than to specify every pixel for an object mask.
However, we found that current interactive VOS methods require multiple rounds to refine the results, which impedes their efficiency in real-world applications.

\begin{figure}[t]
\centering
\includegraphics[width=\linewidth]{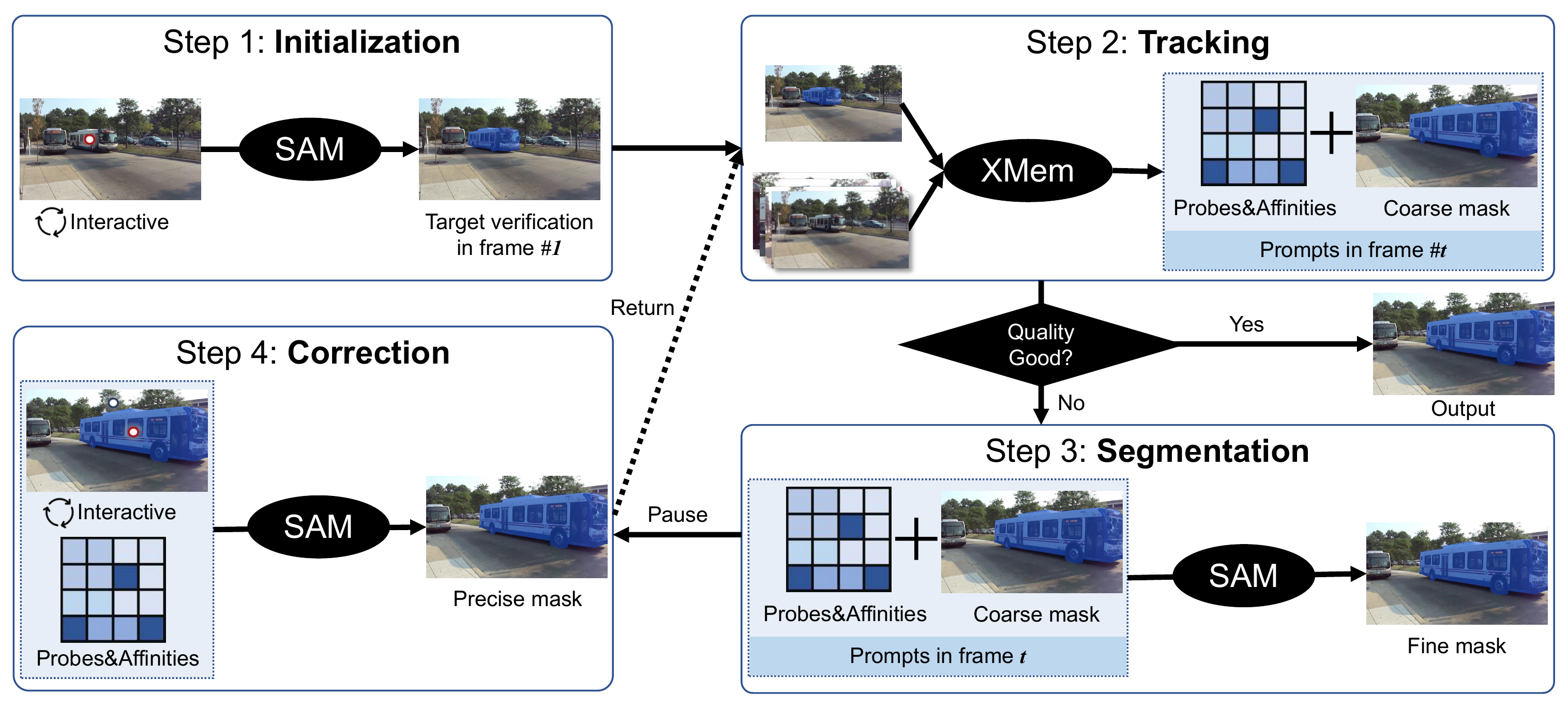}
\caption{Pipeline of our proposed Track Anything Model (TAM). Only within one round of inference can the TAM obtain impressive tracking and segmentation performance on the human-selected target.}
\label{fig:overview}
\end{figure}

\begin{table}
  \caption{Results on DAVIS-2016-val and DAVIS-2017-test-dev datasets~\cite{davis}.}
  \label{davis1617}
  \centering
   \small
   \setlength\tabcolsep{4pt}
  \begin{tabular}{l|c|c|c|ccc|ccc}
    \toprule
    & & & &\multicolumn{3}{c|}{DAVIS-2016-val} &\multicolumn{3}{c}{DAVIS-2017-test-dev} \\
    Method & Venue & Initialization & Evaluation& $J\&F$ & $J$ &$F$ &$J\&F$ & $J$ &$F$\\
    \midrule
    STM~\cite{stm} & ICCV2019 &Mask & One Pass &89.3 &88.7 &89.9 & 72.2 & 69.3 & 75.2 \\
    AOT~\cite{aot} &NeurIPS2021 &Mask & One Pass & 91.1 & 90.1 & 92.1 &  79.6 & 75.9 & 83.3 \\
    XMem~\cite{xmem} & NeurIPS2022 &Mask & One Pass & 92.0 &90.7 &93.2 &  81.2 & 77.6 & 84.7\\
    \midrule
    SiamMask~\cite{siammask}& CVPR2019 &Box & One Pass & 69.8 &71.7 &67.8 &- &- &- \\
    \midrule
    MiVOS~\cite{mivos} & CVPR2021 &Scribble &8 Rounds &91.0 &89.6 &92.4 &78.6 &74.9 &82.2\\
    \midrule
    TAM (Proposed) &- & Click & One Pass & 88.4 & 87.5 &89.4 & 73.1 & 69.8 & 76.4\\
    \bottomrule
  \end{tabular}
\end{table}

\subsection{Implementation}\label{implementation}

Inspired by SAM, we consider tracking anything in videos.
We aim to define this task with high interactivity and ease of use.
It leads to ease of use and is able to obtain high performance with very little human interaction effort.
Figure~\ref{fig:overview} shows the pipeline of our Track Anything Model (TAM).
As shown, we divide our Track-Anything process into the following four steps:

\textbf{Step 1: Initialization with SAM~\cite{sam}.}
As SAM provides us an opportunity to segment a region of interest with weak prompts, \textit{e.g.}, points, and bounding boxes, we use it to give an initial mask of the target object.
Following SAM, users can get a mask description of the interested object by a click or modify the object mask with several clicks to get a satisfactory initialization.

\textbf{Step 2: Tracking with XMem~\cite{xmem}.}
Given the initialized mask, XMem performs semi-supervised VOS on the following frames.
Since XMem is an advanced VOS method that can output satisfactory results on simple scenarios, we output the predicted masks of XMem on most occasions.
When the mask quality is not such good, we save the XMem predictions and corresponding intermediate parameters, \textit{i.e.}, probes and affinities, and skip to step 3.

\textbf{Step 3: Refinement with SAM~\cite{sam}.}
We notice that during the inference of VOS models, keep predicting consistent and precise masks are challenging.
In fact, most state-of-the-art VOS models tend to segment more and more coarsely over time during inference.
Therefore, we utilize SAM to refine the masks predicted by XMem when its quality assessment is not satisfactory.
Specifically, we project the probes and affinities to be point prompts for SAM, and the predicted mask from Step 2 is used as a mask prompt for SAM.
Then, with these prompts, SAM is able to produce a refined segmentation mask.
Such refined masks will also be added to the temporal correspondence of XMem to refine all subsequent object discrimination.

\textbf{Step 4: Correction with human participation.}
After the above three steps, the TAM can now successfully solve some common challenges and predict segmentation masks.
However, we notice that it is still difficult to accurately distinguish the objects in some extremely challenging scenarios, especially when processing long videos.
Therefore, we propose to add human correction during inference, which can bring a qualitative leap in performance with only very small human efforts.
In detail, users can compulsively stop the TAM process and correct the mask of the current frame with positive and negative clicks.

\section{Experiments}

\subsection{Quantitative Results}

To evaluate TAM, we utilize the validation set of DAVIS-2016 and test-development set of DAVIS-2017~\cite{davis}.
Then, we execute the proposed TAM as demonstrated in Section~\ref{implementation}.
The results are given in Table~\ref{davis1617}.
As shown, our TAM obtains $J\&F$ scores of 88.4 and 73.1 on DAVIS-2016-val and DAVIS-2017-test-dev datasets, respectively.
Note that TAM is initialized by clicks and evaluated in one pass.
Notably, we found that TAM performs well when against difficult and complex scenarios.



\begin{figure}[t]
\centering
\includegraphics[width=\linewidth]{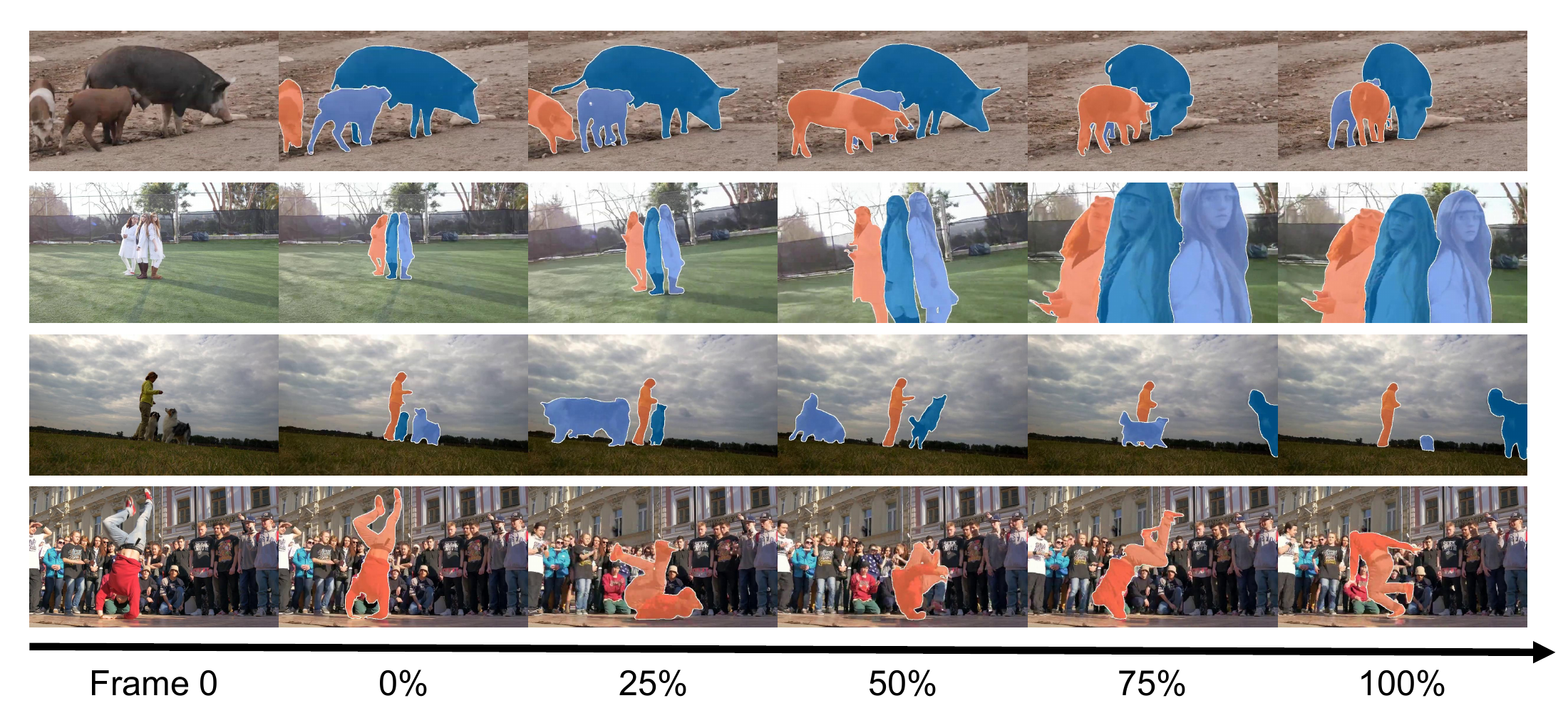}
\caption{Qualitative results on video sequences from DAVIS-16 and DAVIS-17 datasets~\cite{davis}.}
\label{fig:davisresult}
\end{figure}

\begin{figure}[t]
\centering
\includegraphics[width=\linewidth]{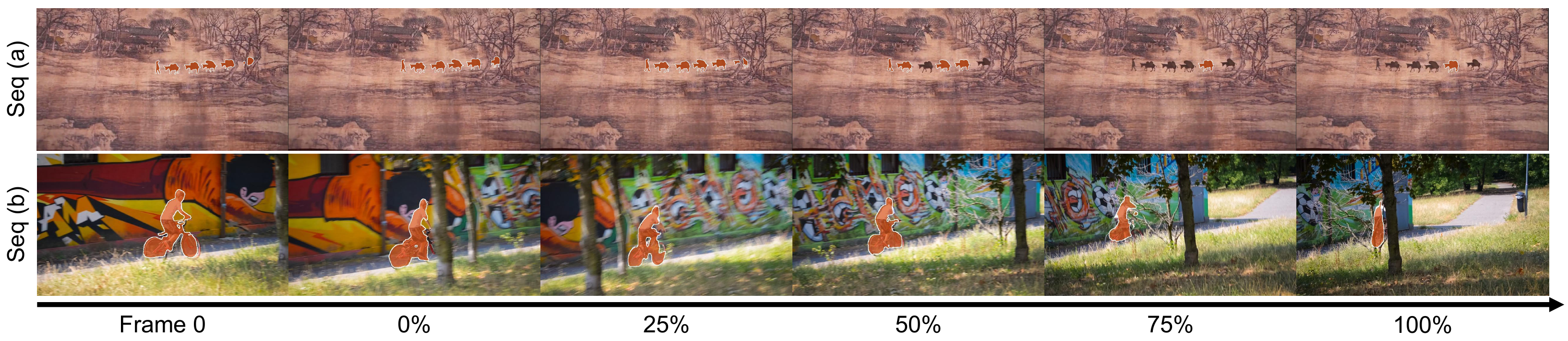}
\caption{Failed cases.}
\label{fig:failedcases}
\end{figure}

\subsection{Qualitative Results}

We also give some qualitative results in Figure~\ref{fig:davisresult}.
As shown, TAM can handle multi-object separation, target deformation, scale change, and camera motion well, which demonstrates its superior tracking and segmentation abilities within only click initialization and one-round inference.

\subsection{Failed Cases}
We here also analyze the failed cases, as shown in Figure~\ref{fig:failedcases}.
Overall, we notice that the failed cases typically appear on the following two occasions.
1)
Current VOS models are mostly designed for short videos, which focus more on maintaining short-term memory rather than long-term memory.
This leads to mask shrinkage or lacking refinement in long-term videos, as shown in seq (a).
Essentially, we aim to solve them in step 3 by the refinement ability of SAM, while its effectiveness is lower than expected in realistic applications.
It indicates that the ability of SAM refinement based on multiple prompts can be further improved in the future.
On the other hand, human participation/interaction in TAM can be an approach to solving such difficulties, while too much interaction will also result in low efficiency.
Thus, the mechanism of long-term memory preserving and transient memory updating is still important.
2) When the object structure is complex, \textit{e.g.}, the bicycle wheels in seq (b) contain many cavities in groundtruth masks. We found it very difficult to get a fine-grained initialized mask by propagating the clicks.
Thus, the coarse initialized masks may have side effects on the subsequent frames and lead to poor predictions.
This also inspires us that SAM is still struggling with complex and precision structures.

\begin{figure}[t]
\centering
\includegraphics[width=\linewidth]{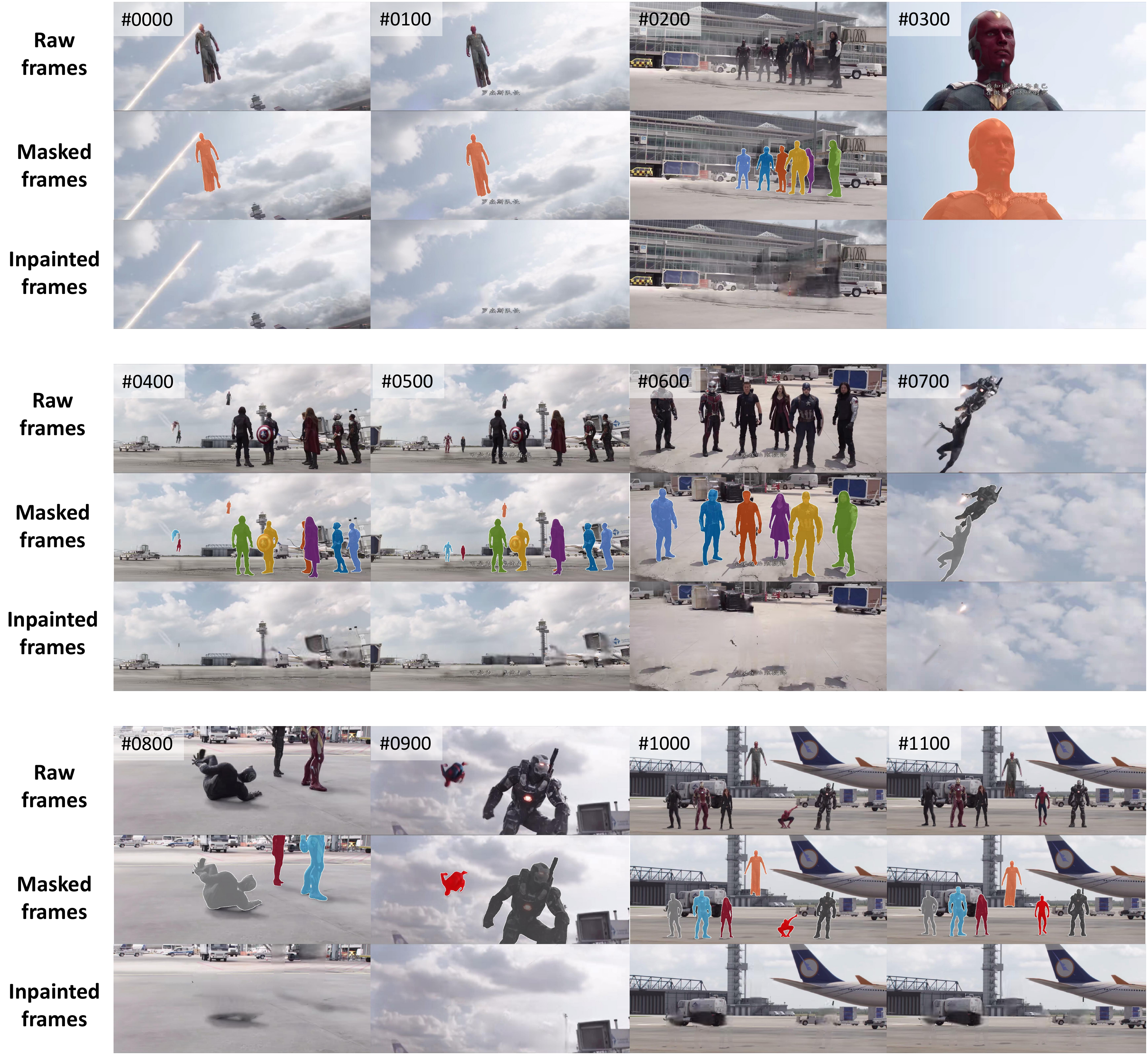}
\caption{Raw frames, object masks, and inpainted results from the movie \textit{Captain America: Civil War (2016)}.}
\label{fig:captain}
\end{figure}

\section{Applications}
The proposed Track Anything Model (TAM) provides many possibilities for flexible tracking and segmentation in videos.
Here, we demonstrate several applications enabled by our proposed method.
In such an interactive way, diverse downstream tasks can be easily achieved.

\textbf{Efficient video annotation.}
TAM has the ability to segment the regions of interest in videos and flexibly choose the objects users want to track. Thus, it can be used for video annotation for tasks like video object tracking and video object segmentation.
On the other hand, click-based interaction makes it easy to use, and the annotation process is of high efficiency.

\textbf{Long-term object tracking.}
The study of long-term tracking is gaining more and more attention because it is much closer to practical applications.
Current long-term object tracking task requires the tracker to have the ability to handle target disappearance and reappearance while it is still limited in the scope of trimmed videos.
Our TAM is more advanced in real-world applications which can handle the shot changes in long videos.

\textbf{User-friendly video editing.}
Track Anything Model provides us the opportunities to segment objects
With the object segmentation masks provided by TAM, we are then able to remove or alter any of the existing objects in a given video.
Here we combine E$^2$FGVI~\cite{e2fgvi} to evaluate its application value.

\textbf{Visualized development toolkit for video tasks.}
For ease of use, we also provide visualized interfaces for multiple video tasks, \textit{e.g.}, VOS, VOT, video inpainting, and so on.
With the provided toolkit, users can apply their models on real-world videos and visualize the results instantaneously.
Corresponding demos are available in Hugging Face\footnote{\url{https://huggingface.co/spaces/watchtowerss/Track-Anything}}.

To show the effectiveness, we give a comprehensive test by applying TAM on the movie \textit{Captain America: Civil War (2016)}.
Some representative results are given in Figure \ref{fig:captain}.
As shown, TAM can present multiple object tracking precisely in videos with lots of shot changes and can further be helpful in video inpainting.






\bibliographystyle{plain}
\bibliography{neurips_2022}

\end{document}